\documentclass[letterpaper, 10 pt, conference]{ieeeconf}  

\IEEEoverridecommandlockouts                              

\usepackage{epsfig} 
\usepackage{mathptmx} 
\usepackage{times} 
\usepackage{amsmath} 
\usepackage{amssymb}  

\usepackage{graphics}
\usepackage[utf8]{inputenc}
\usepackage{graphicx,subfigure}
\usepackage{booktabs}
\usepackage{amsmath} 
\usepackage{amssymb}  
\usepackage{amsfonts}
\usepackage{mathtools}
\usepackage[font=small]{caption}
\usepackage{booktabs}
\usepackage{wrapfig}
\usepackage{flushend}
\usepackage{pifont}
\usepackage[hidelinks]{hyperref}
\usepackage[capitalise]{cleveref}
\usepackage{bbm}
\usepackage{tikz}
\usepackage{bbold}
\usepackage{adjustbox}
\def\BibTeX{{\rm B\kern-.05em{\sc i\kern-.025em b}\kern-.08em
    T\kern-.1667em\lower.7ex\hbox{E}\kern-.125emX}}

\usepackage{siunitx}
\usepackage{bm}

\DeclareMathAlphabet\mathbfcal{OMS}{cmsy}{b}{n}

\newcommand{\orangestar}{{\color{orange}$\bigstar$}}
\newcommand{\reddot}{{\color{red}$\bullet$}}
\newcommand{\blacksq}{$\blacksquare$}

\usepackage{pifont}
\newcommand{\greentriangle}{{\color{green}\ding{116}}} 
\newcommand{\redanchor}{{\color{red}\ding{108}}}

\newcommand{\cyansquare}{{\color{cyan}\ding{110}}}
\newcommand{\greycircle}{{\color{gray}\ding{108}}}
\newcommand{\graysquare}{{\color{gray}\ding{110}}}
\newcommand{\redsq}{{\color{red}\ding{110}}}
\newcommand{\bluesq}{{\color{blue}\ding{110}}}
\definecolor{semigray}{gray}{0.7}
\newcommand{\semigraysquare}{{\color{semigray}\ding{110}}}
\definecolor{lightblue}{rgb}{0.53, 0.81, 0.98}
\newcommand{\bluecircle}{{\color{lightblue}\ding{108}}}
\DeclarePairedDelimiterX{\infdivx}[2]{(}{)}{%
  #1\;\delimsize|\delimsize|\;#2%
}

\usepackage{booktabs}
\usepackage[style=ieee, url=false, doi=false, natbib=true, mincitenames=1, maxcitenames=1]{biblatex}

\title{\LARGE \bf
Scene Informer: Anchor-based Occlusion Inference and Trajectory Prediction in Partially Observable Environments}

\author{Bernard Lange$^{1}$, Jiachen Li$^{2}$, and Mykel J.~Kochenderfer$^{1}$
\thanks{$^{1}$Stanford Intelligent Systems Laboratory,
        Stanford University, Stanford, CA 94305, USA
        \texttt{\{blange, jiachen\_li, mykel\}@stanford.edu}}%
\thanks{$^{2}$Trustworthy Autonomous Systems Laboratory,
        University of California, Riverside, Riverside, CA 92507, USA
        \texttt{jiachen.li@ucr.edu}}%
}

\begin{document}

\maketitle
\thispagestyle{empty}
\pagestyle{empty}

\begin{abstract}
Navigating complex and dynamic environments requires autonomous vehicles (AVs) to reason about both visible and occluded regions. This involves predicting the future motion of observed agents, inferring occluded ones, and modeling their interactions based on vectorized scene representations of the partially observable environment. However, prior work on occlusion inference and trajectory prediction have developed in isolation, with the former based on simplified rasterized methods and the latter assuming full environment observability. We introduce the Scene Informer, a unified approach for predicting both observed agent trajectories and inferring occlusions in a partially observable setting. It uses a transformer to aggregate various input modalities and facilitate selective queries on occlusions that might intersect with the AV's planned path. The framework estimates occupancy probabilities and likely trajectories for occlusions, as well as forecast motion for observed agents. We explore common observability assumptions in both domains and their performance impact. Our approach outperforms existing methods in both occupancy prediction and trajectory prediction in partially observable setting on the Waymo Open Motion Dataset. Our implementation with additional visualizations is available at \href{https://github.com/sisl/SceneInformer}{https://github.com/sisl/SceneInformer}.

\end{abstract}


\section{INTRODUCTION}
Safe navigation through dynamic environments necessitates reasoning about both occluded and visible parts of the environment. Traffic participants analyze the social cues from other agents and the topography of the scene to infer hypothetical future scenarios. For instance, a vehicle slowing down near a crosswalk could imply a pedestrian is about to cross. Experienced drivers incorporate the uncertainty tied to potential states of the occluded environment into their decision-making to enhance maneuver safety. Similar reasoning is crucial in safety-critical and dynamic robotic applications, such as self-driving cars. This paper aims to tackle the task of occlusion inference and trajectory prediction in autonomous vehicles (AVs).

In the context of AVs, a suite of sensors delivers an environment representation to the perception and sensor fusion modules. These modules generate representations used in downstream tasks such as trajectory prediction and planning. Naturally, these sensors are sometimes partially occluded, leading to incomplete representations and erroneous outcomes from the downstream tasks. Hence, occlusion awareness is critical to safe driving. Two primary methods can integrate occlusion reasoning into an AV: occlusion-aware decision making~\cite{bouton2018scalable, zhang2018unreasonable, lin2019decision, wray2021pomdps, mun2022occlusion, christianos2022planning}, which involves planning conservatively around occluded areas, and occlusion inference~\cite{afolabi2018people, itkinaicra2022, krueger2022recognition} with a focus on deducing the presence of objects in the occlusions. This work is focused on occlusion inference, due to its ability to offer an explicit representation of inferred occluded agents for downstream tasks, and its integration with trajectory prediction of observed agents.

\begin{figure}[t]
  \centering
  \centerline{\includegraphics[width=1.0\columnwidth]{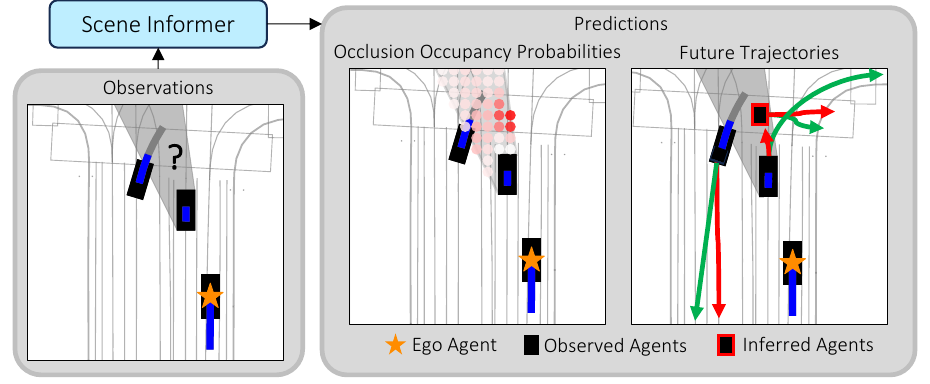}}
  \caption{We introduce Scene Informer, an end-to-end prediction framework that considers both observed and occluded agents in a partially observable environment. It forecasts multi-modal futures for observed agents and estimates occupancy probabilities and most likely trajectories originating from the occlusion.}
  \label{fig:overview}
  \vspace{-2em}
\end{figure}

Effective reasoning about a partially observable environment necessitates considering the observed and occluded agents, their future motion, and their interactions with one another. This involves processing vectorized inputs from perception frameworks, such as agent trajectories, agent properties, and environmental maps.
Despite these requirements, occlusion inference and trajectory prediction have each been developed in isolation, with limited integration between the two domains. Prior work on occlusion inference has primarily operated under simplified settings. They reason in terms of fixed-size grids with occupancy probabilities of agents and do not leverage the diverse input modalities available to modern AV~\cite{afolabi2018people, itkinaicra2022, krueger2022recognition, christianos2022planning}. Furthermore, they do not model the interaction between predicted occluded and visible agents. On the other hand, trajectory prediction approaches that predict the motion of observed agents typically assume full observability of the environment~\cite{chai2019multipath, ivanovic2020mats, gu2021densetnt, shi2022motion}.

To rectify these limitations, we introduce an end-to-end, learning-based framework for environment prediction in partially observable settings called \textit{Scene Informer}. Our proposed method is designed to infer any occlusion of interest and predict the motion of observed agents. It uses a transformer to aggregate an arbitrary number of input modalities that include histories and properties of observed agents along with a lane graph. It allows selective inference of occluded areas, enabling queries for occlusions that might interact with the AV's planned trajectory. It infers a set of occupancy probabilities along with the most likely trajectories for each occlusion, and future trajectories for the observed agents, as shown in \cref{fig:overview}. 
Our framework is the first end-to-end comprehensive environment prediction solution that both infers occlusions and predicts observed agent trajectories. 


We evaluate our framework on the Waymo Open Motion Dataset (WOMD)~\cite{ettinger2021large} by simulating occlusions with a line-of-sight method from the ego vehicle perspective. Our method is compared with recent work on occlusion inference~\cite{afolabi2018people,itkinaicra2022}, fully observable trajectory prediction, and simplified variations of Scene Informer. We investigate various observability assumptions prevalent in both occlusion inference and trajectory prediction frameworks, and assess the impact of end-to-end training on overall performance. Our proposed framework achieves state-of-the-art performance in terms of both occupancy prediction and trajectory prediction of occluded objects, and demonstrates increased robustness to partial observability when forecasting observed agents. In summary, our contributions are as follows:


\begin{itemize}
    \item We propose a novel occlusion inference framework called Scene Informer that uses a transformer to infer selected obstructed parts of the scene, and predict future trajectories of the observed agents.
    \item We investigate the influence of different observability assumptions commonly used in the occlusion inference and trajectory prediction frameworks and their impact on the final performance.
    \item We demonstrate the superior performance of our approach on the Waymo Open Motion Dataset (WOMD) compared to prior work on occlusion inference and fully observable trajectory prediction.
\end{itemize}
 \begin{figure*}[htbp]
   \vspace{1.0em}
  \centering
  \centerline{\includegraphics[width=0.85\textwidth]{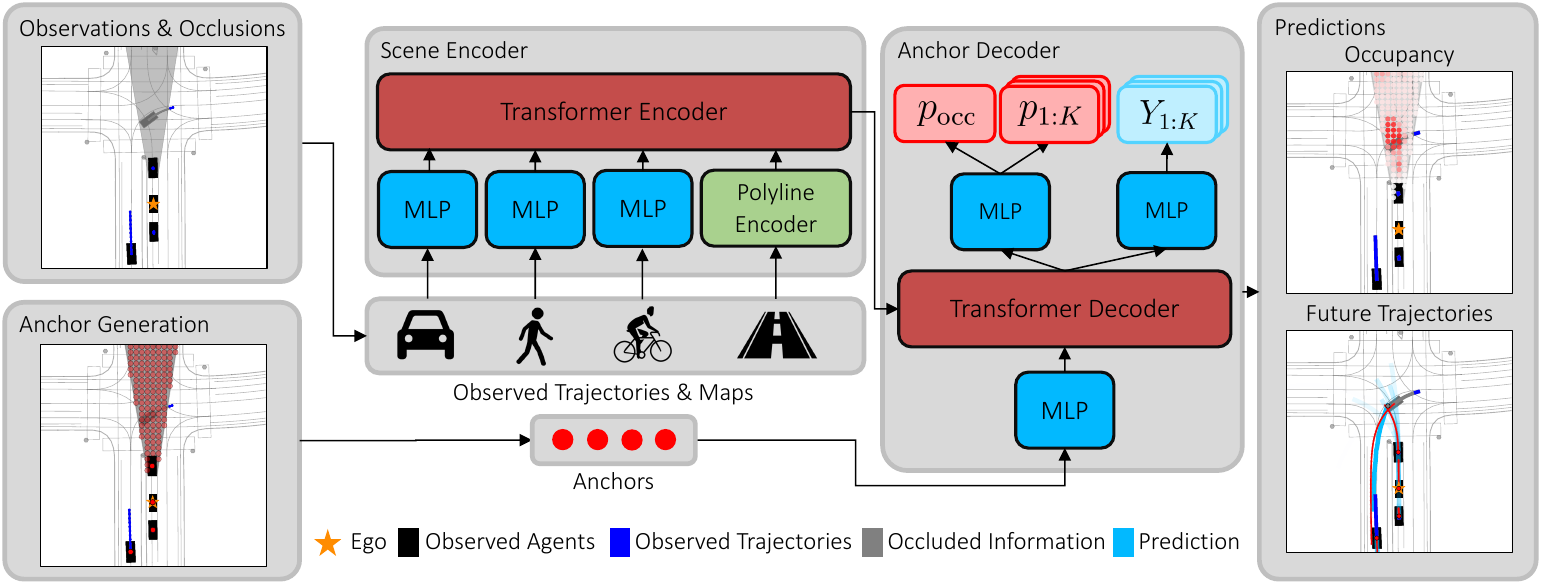}} 
  \caption{Scene Informer consists of a scene encoder and anchor decoder. It reasons in terms of anchors (\redanchor) that are assigned to each observed agent and randomly populated in the occlusion of interest. Scene encoder aggregates different observation input modalities and creates scene embeddings. Anchor decoder cross-attends between scene embeddings and anchors, and outputs predicted occupancy probability $p_{occ}$ and $K$ most likely future trajectories $Y_{1:K}$ (\cyansquare) with the probability of each trajectory $p_{1:K}$ for each anchor.}
  \label{fig:detailed_overview}
  \vspace{-2.0em}
\end{figure*}

\section{RELATED WORK} 
We discuss the prior work on occlusion inference, occlusion-aware decision making, and trajectory prediction.

\textbf{Occlusion Inference.} 
Prior work uses interactions between observed agents in the scene to infer the obstructed parts of the environment. \citet{afolabi2018people} clusters observed interactions between human drivers and crosswalks to learn occupancy grid maps (OGMs) of the occlusions. \citet{hara2020predicting} uses camera observations to infer the state of the blindspot. \citet{itkinaicra2022} accounts for the multimodality of the scene by learning a driver sensor model with a conditional variational autoencoder (CVAE)~\citep{sohn2015learning} which maps observed trajectories to a latent vector. 
Then they fuse multiple maps acquired from different observed agents with evidential theory~\citep{dempster1968generalization} to generate the OGM of the occluded area. \citet{christianos2022planning} extends it by proposing a two-stage training procedure with an additional CVAE that predicts future trajectories of inferred agents. Prior work has relied on fixed-size OGMs and rasterized networks and is not trained end-to-end. They fail to capture the semantics of the scenes and attempt to infer all occlusions. We propose a single-stage architecture that uses vectorized representations to infer any occlusion in terms of occupancy and likely future trajectories, and forecast trajectories of observed agents.


\textbf{Occlusion Aware Decision Making.} 
The planner can incorporate sensor occlusions in the observation model~\citep{bouton2018scalable, mun2022occlusion}, or include imaginary agents in the occlusions often following some worst-case scenario analysis~\citep{chae2020virtual, wray2021pomdps, hoermann2017entering, hubmann2019pomdp}. \citet{bouton2018scalable} demonstrates that a POMDP policy can safely navigate with sensor occlusions. \citet{chae2020virtual} incorporates imaginary occluded objects into the overtaking controller. \citet{wray2021pomdps} includes virtual agents outside the vehicle's field of view in the observation model of a T-intersection POMDP policy. \citet{hoermann2017entering} aggregates both object-based and object-free representations (e.g. occupancy grid maps) to safely navigate left turns. \citet{hubmann2019pomdp} generalizes phantom vehicle reasoning to any urban scenarios. \citet{nager2019lies} extends it to different classes of virtual agents. \citet{mun2022occlusion} encodes occluded observations with a variational model~\citep{kingma2013auto} and trains an agent to navigate crowded scenes. \citet{christianos2022planning} predicts likely occluded trajectories for the planner.

\textbf{Fully Observable Trajectory Prediction.} Trajectory prediction forecasts the motion of agents based on the assumption of full observability. The effectiveness of the approach is determined by the methodology used to aggregate the input modalities. In rasterized approaches \citep{lee2017desire, casas2018intentnet, hong2019rules, cui2019multimodal, phan2020covernet, choi2021shared}, all modalities are represented with image-like representations and encoded with convolutional network. An alternative way is to represent different input modalities as a set and process them with recurrent neural networks \citep{salzmann2020trajectron++, mercat2020multi, ma2021continual} or graph neural networks \citep{casas2020spagnn, cao2021spectral, khandelwal2020if, li2020evolvegraph, gu2021densetnt, girase2021loki, gao2020vectornet, li2021spatio, zhou2022grouptron}. Recently, transformer-based approaches \citep{ngiam2021scene, nayakanti2022wayformer, shi2022motion} have been achieving state-of-the-art results on numerous benchmarks. In our proposed approach, we adopt a transformer-based framework to infer occlusions and predict the motion of observed agents addressing the unrealistic full observability assumption of prior methods.

\section{Scene Informer}
Reasoning over observed and occluded agents that might interfere with a planned trajectory is critical to ensure the safety of the AV maneuvers. It involves processing a range of vectorized input modalities provided by upstream perception frameworks, such as observed agents and lane graphs while taking into account the partial observability of the environment. 
We propose an environment prediction framework called \textit{Scene Informer} (see \cref{fig:detailed_overview}) that takes in observed trajectories of other agents, vectorized maps, and infers the true state of the occlusion interest and the future motion of observed agents. Our framework consists of transformer encoder and decoder pairing. The encoder takes in vectorized representation of the observed agents and maps to create the scene embeddings~\cite{ngiam2021scene, nayakanti2022wayformer}. 
The decoder receives the scene embeddings and reasons in terms of spatial anchors for both observed agents and occlusions. For observed agents, spatial anchors indicate the most recent position. In the case of occlusions, they represent points randomly placed within the occlusion of interest. It allows intelligent queries for occlusions that might interact with the planned trajectory rather than inferring all occlusions at once, which is computationally expensive and unnecessary. Finally, for each provided anchor, the decoder infers a probability of occupancy and the $k$ most likely trajectories with their corresponding probabilities to model the multi-future behavior of the observed and potentially present but occluded agents. 
 
\subsection{Problem Statement and Input Representation}
\label{sec:problem_statement}
Given an environment state representation $S_{\text{past}}$ that consists of observed agent information $O_{\text{past}}$ and road map information $M$, our objective is to infer the occlusion and predict future trajectories from the perspective of the ego vehicle. Agent information $O_{\text{past}} \in \mathbb{R}^{N_o \times H \times D_a}$ represents $N_o$ agents over history horizon $H$ with $D_a$ features consisting of position, heading, velocity, and agent's dimensions. Road map information is denoted as $M \in \mathbb{R}^{N_m \times N_p \times D_m}$ which describes $N_m$ polylines with $N_p$ points with feature vector $D_m$. Each point is represented by its position, vector to the next point in a polyline, and polyline type (e.g. crosswalk, lane boundary, stop sign, etc.) We represent the occlusion to infer and agents to predict by a set of spatial anchors $A \in \mathbb{R}^{N_a \times 2}$, where $N_a$ is the number of anchors defined by their position. $N_o$ anchors are assigned to the most recent positions of the observed agents, while $N_a - N_o$ anchors are uniformly distributed within the occluded areas, as shown in \cref{fig:detailed_overview}.
For each anchor, we predict $k$ most likely trajectories $Y$ over the prediction horizon $P$ with probability distribution $p_{1:K}$, and probability of the anchor being occupied $p_{\text{occ}}$. 
\subsection{Framework}
\label{sec:occlusion_informer}
We use a transformer architecture that has been shown to excel at modeling multimodal and unstructured inputs of varying sizes~\cite{ngiam2021scene, nayakanti2022wayformer, shi2022motion}. Our proposed framework consists of \textit{scene encoder} and \textit{anchor decoder}. The scene encoder processes multimodal input, such as past agent trajectories and vectorized maps, and generates scene embeddings. The anchors that define agent and occlusions to predict are provided to the anchor decoder which cross-attends them with the scene embeddings to generate anchor embeddings. Those embeddings are then used to predict the probability of the anchor being occupied, and top $k$ future trajectories with their corresponding probability distribution. 

\textbf{Scene encoder:} It aggregates the state representation $S_{\text{past}}$ that includes agent information $O_{\text{past}}$ and road map information $M$ modalities. Each agent type is encoded with a dedicated multi-layer perceptron, and each road polyline is modeled with a PointNet polyline encoder layer~\citep{qi2017pointnet, shi2022motion} as it allows us to encode polylines of varying lengths:
\begin{equation}
M_{\text{enc}},\; O_{\text{enc}} = \phi(\text{MLP}(M)),\; \text{MLP}(O_{\text{past}}) 
\end{equation}
where MLP$(\cdot)$ is a multi-layer perceptron network, and $\phi$ is a max-pooling operator over the last feature dimension. For different agent types, we have a dedicated set of parameters for MLP. Subsequently, all representations are concatenated to form $S_{\text{in}} = [O_{\text{enc}}, M_{\text{enc}}] \in \mathbb{R}^{(N_o + N_m) \times D_{\text{enc}}}$. They are provided to the transformer encoder which aggregates input modalities with a varying number of tokens:
\begin{equation}
    S_{\text{\text{emb}}} = \text{TransformerEncoder}(S_{\text{in}})
\end{equation}
where $S_{\text{\text{emb}}} \in \mathbb{R}^{(N_o + N_m) \times D_{\text{out}}}$ are scene embeddings.

\textbf{Anchor decoder:} We define anchors $A \in \mathbb{R}^{N_a \times 2}$. Each anchor is projected to a feature dimension $D_{\text{enc}}$ consistent with other encodings. It is then provided to the transformer decoder which cross-attends them with other anchors and scene embeddings from the scene encoder:
\begin{equation}
    A_{\text{\text{emb}}} = \text{TransformerDecoder}(\text{MLP}(A), S_{\text{\text{emb}}})
\end{equation}
where $A_{\text{\text{emb}}} \in \mathbb{R}^{N_a \times D}$ are anchor embeddings. The embedding of each anchor is provided to MLP to predict the probability of occupancy $p_{\text{occ}}$, $K$ most likely future trajectories represented with Gaussian Mixture Model $Y \in \mathbb{R}^{K \times P \times 5}$, and a probability distribution over trajectories $p_{1:K}$. Each time step of the predicted trajectory $Y$ is represented with parameters $\left(\mu_{x}, \mu_{y}, \sigma_{x}, \sigma_{y}, \rho \right)$ that define a Gaussian component~\cite{shi2022motion}.


\textbf{Training loss:}
Our loss is a mix of regression, trajectory classification, and anchor occupancy classification components. For a given anchor $i$, if it is occupied, we identify the predicted mode $j$ out of $K$ possible trajectories with a hard-assignments strategy~\cite{chai2019multipath, varadarajan2022multipath++}. Subsequently, we maximize the log-likelihood associated with the selection of mode j, the generation of the ground truth trajectory $GT$ by mode $j$, and the occupation of anchor $i$. Conversely, if anchor $i$ remains unoccupied, the log-likelihood is maximized for its unoccupied state. The cumulative loss is computed by aggregating these components over all anchors: 
\begin{align}
\begin{split}
    \max \mathbb{1}[i = \text{occ.}] \bigg( &\log\Pr(j \mid A_{\text{\text{\text{emb}}},i})
    + \log \Pr(GT \mid Y_{i,j})\bigg) \\ + &\log \Pr(i\mid A_{\text{\text{\text{emb}}},i}).
\end{split} 
\end{align}





\section{EXPERIMENTS}
\label{sec:experiments}
\subsection{Dataset}
We evaluate our framework on the Waymo Open Motion Dataset (WOMD)~\cite{ettinger2021large}. It contains approximately 103k 20s-long scenes recorded at 10Hz and consists of object tracks and vectorized maps. To our best knowledge, it is the only large open-source autonomous driving dataset that contains labels sourced from an offboard perception module. The offboard method is a non-causal approach that is not constrained by real-time performance and uses future observations to detect past objects~\cite{qi2021offboard}. As a result, it offers unparalleled accuracy in tracking objects within occlusions. 
We simulate occlusions with a line-of-sight method in the bird's-eye view (BEV) of an ego-vehicle. While occlusions occur in three dimensions, we investigate the integration of partial observability reasoning into the environment prediction framework which operates in a BEV representation. In our experiments, we explore various degrees of observability:
\begin{itemize}
\item \textit{Full observability}: The true state is fully observed, commonly used in trajectory prediction approaches.
\item \textit{Limited observability}: Only agents visible via line-of-sight from the ego vehicle's perspective are observed.
\item \textit{Partial observability}: Agents visible from the ego vehicle's perspective are observed, along with other randomly selected agents that might be occluded.
\end{itemize}
\begin{figure}[ht]
\tiny
\centering     
\subfigure[Full Obs]{\label{fig:a}\includegraphics[width=30mm, trim={5mm 5mm 5mm 5mm}, clip]{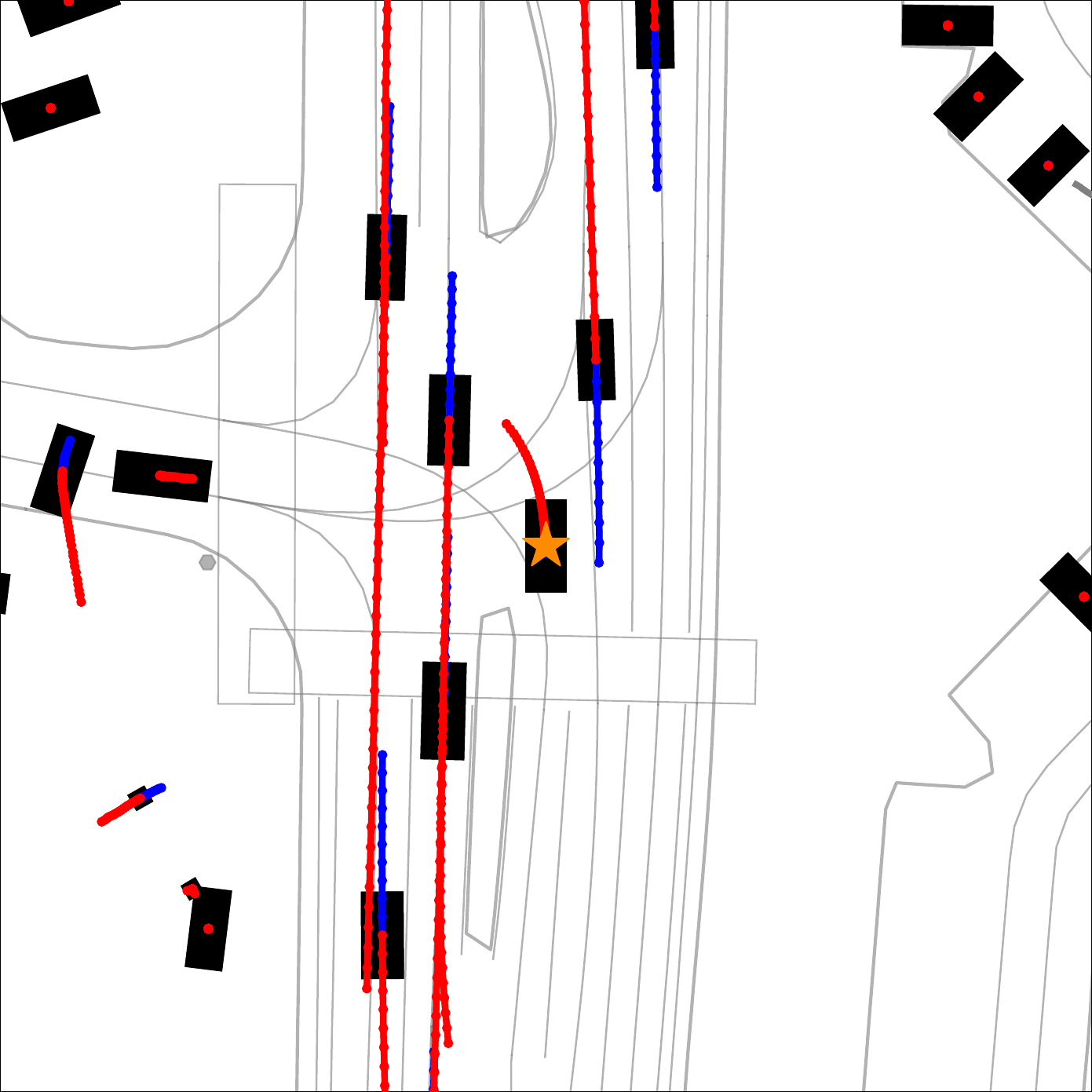}}
\hspace{1.0cm} 
\subfigure[Partial Obs (\SI{50}{\percent})]{\label{fig:a}\includegraphics[width=30mm, trim={5mm 5mm 5mm 5mm}, clip]{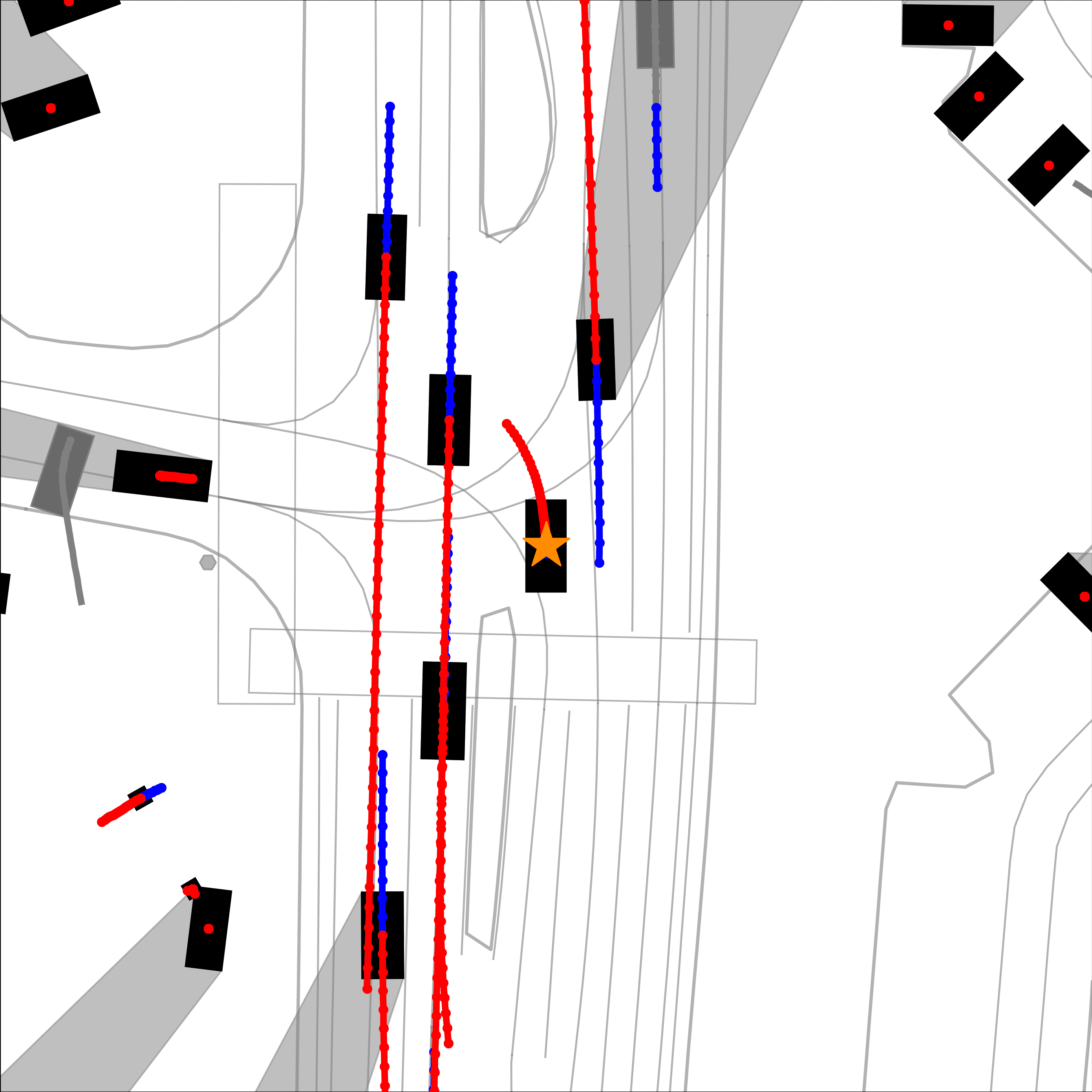}}

\vspace{-0.3cm} 

\subfigure[Limited Obs]{\label{fig:b}\includegraphics[width=30mm, trim={5mm 5mm 5mm 5mm}, clip]{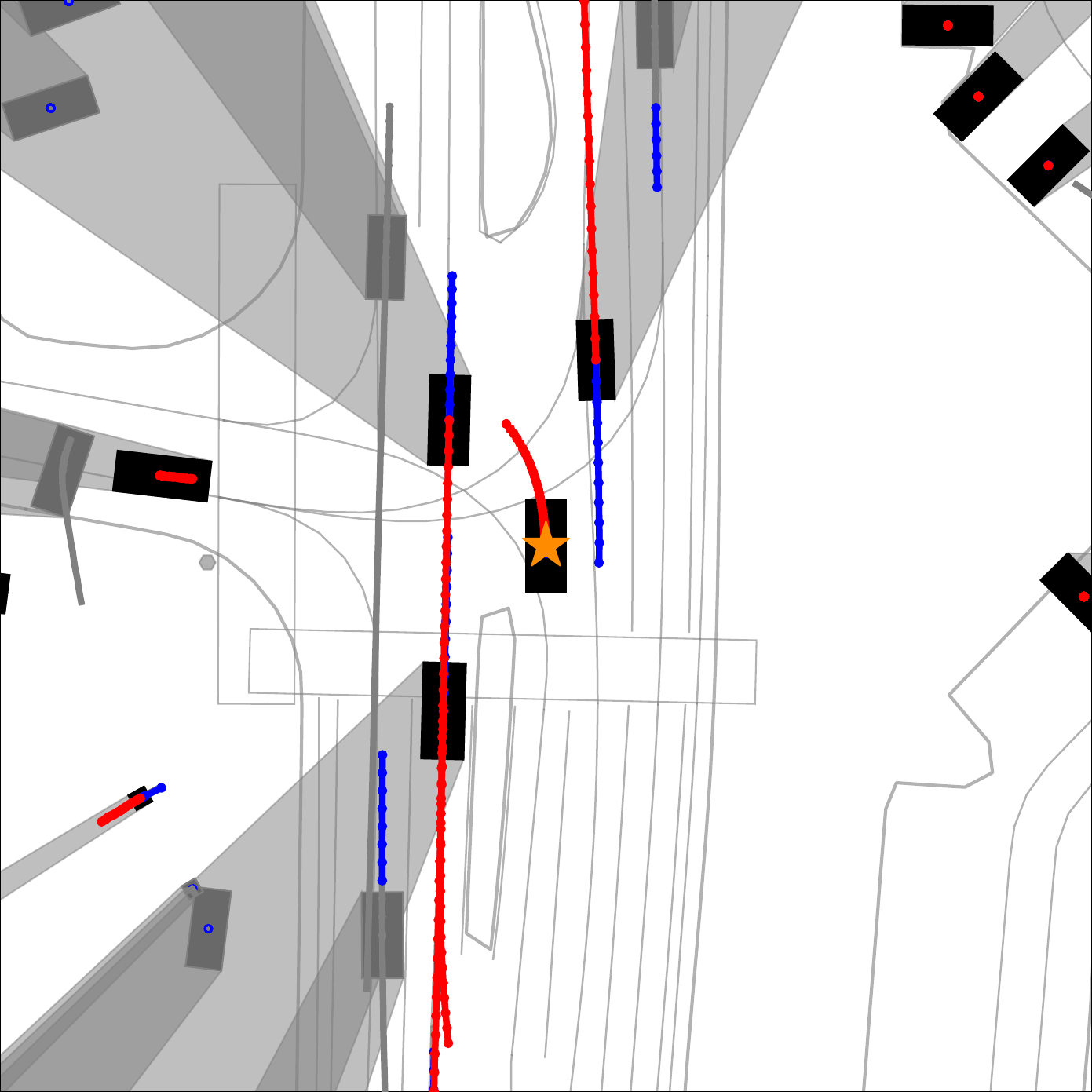}}
\hspace{1.0cm} 
\subfigure[Full Obs (Occ. Inf.)]{\label{fig:b}\includegraphics[width=30mm, trim={5mm 5mm 5mm 5mm}, clip]{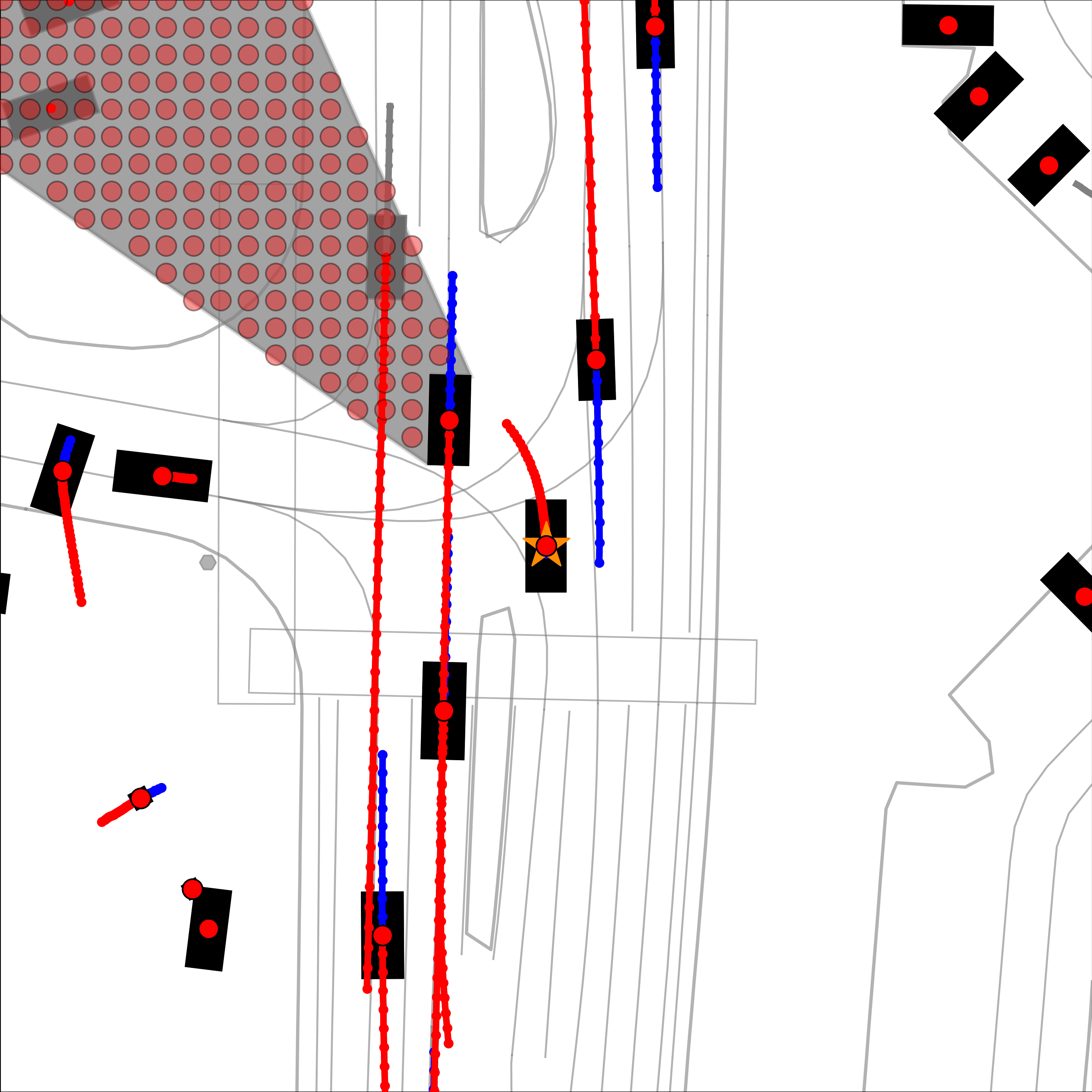}}
\vspace{-0.25cm}

\caption{Visualization of the dataset. Each sample contains agents (\blacksq) with a history of observations (\bluesq) and future trajectory (\redsq) in the frame of the ego vehicle (\orangestar). We explore the following variations: (a) Full Observability. (b) Partial Observability with \SI{50}{\percent} of generating occlusions (\semigraysquare). (c) Limited Observability with all possible occlusions. (d) Full Observability with a single occlusion (used for Scene Informer training). Anchors (\reddot) are assigned to the last time step of observed agents and populated in the occlusion. Occluded agents and observations are grey (\graysquare).} 
\vspace{-0.4cm}
\label{fig:dataset}
\end{figure}

\renewcommand{\thesubfigure}{}
\begin{figure*}[ht]
\tiny
\centering     
\subfigure[]{\label{fig:a}\includegraphics[width=36mm]{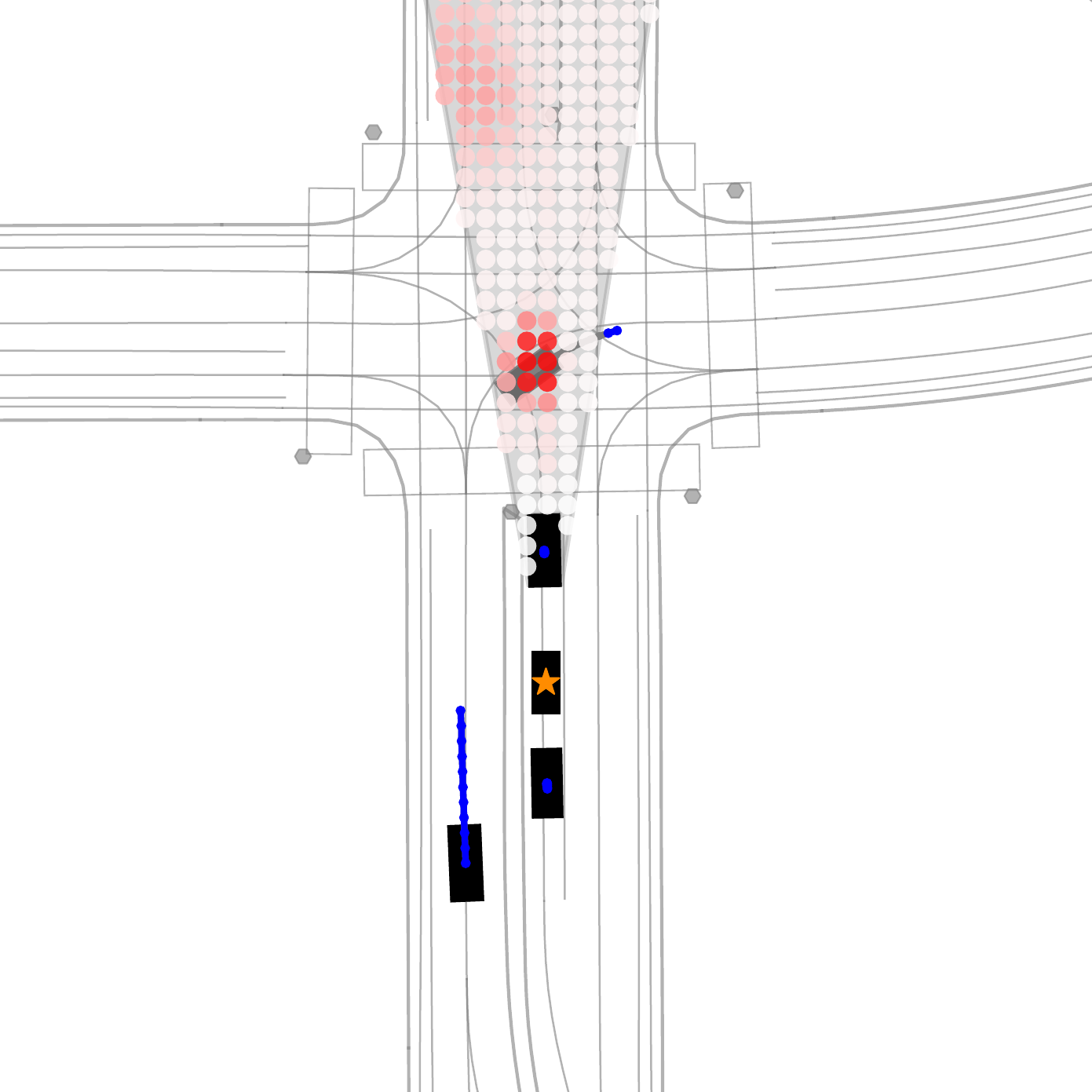}}
\hspace{0.25cm} 
\subfigure[]{\label{fig:b}\includegraphics[width=36mm]{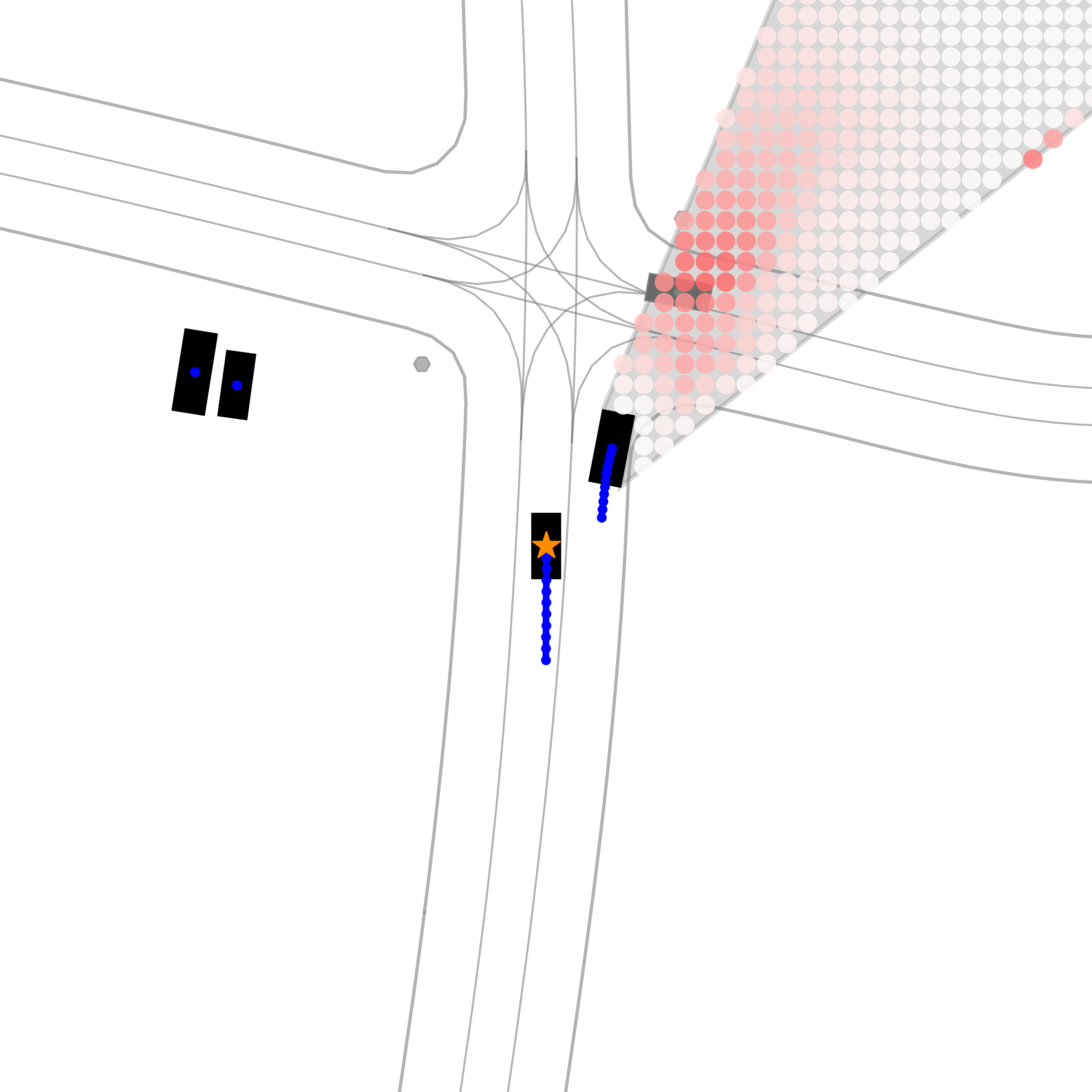}} 
\hspace{0.25cm} 
\subfigure[]{\label{fig:a}\includegraphics[width=36mm]{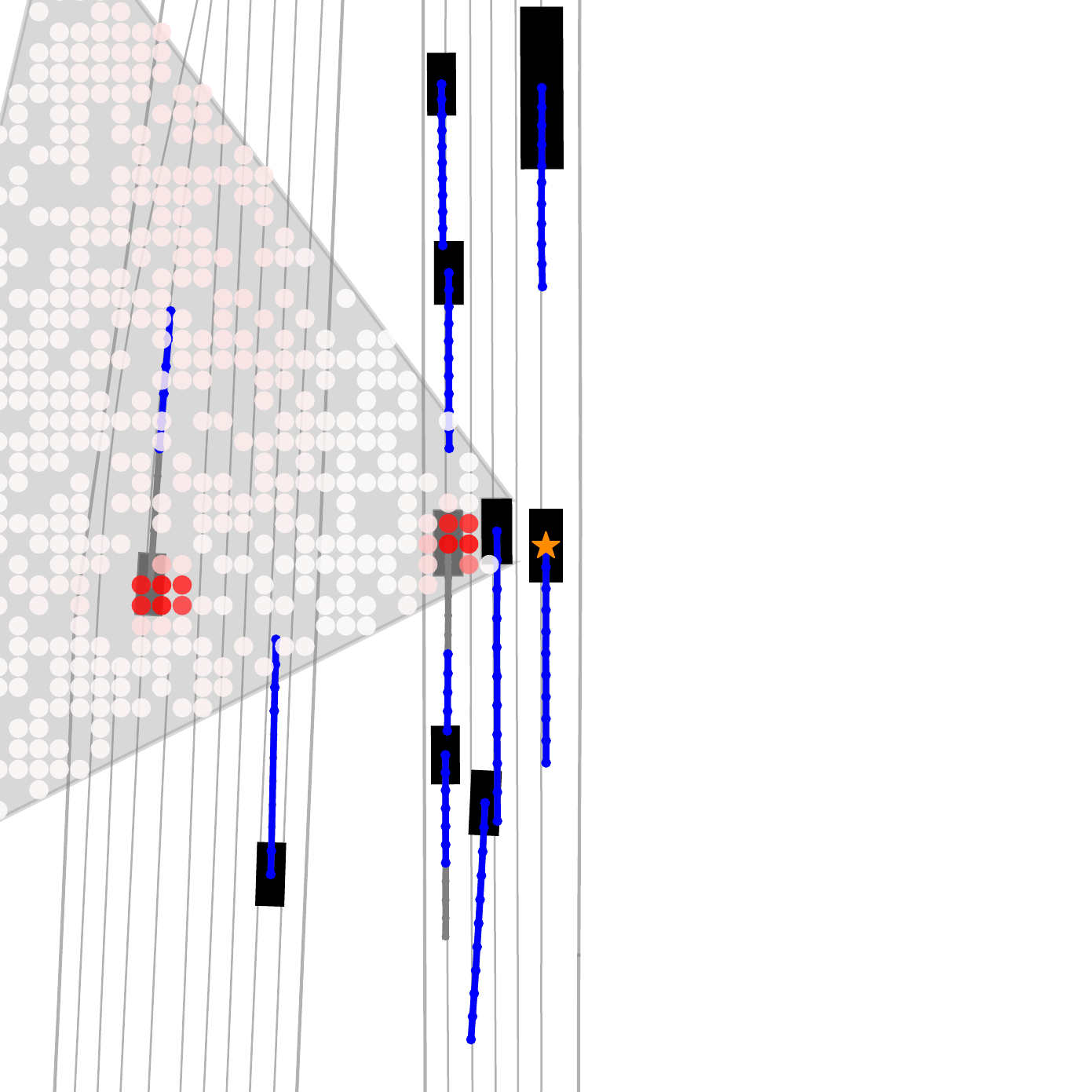}}
\hspace{0.25cm} 
\subfigure[]{\label{fig:b}\includegraphics[width=36mm]{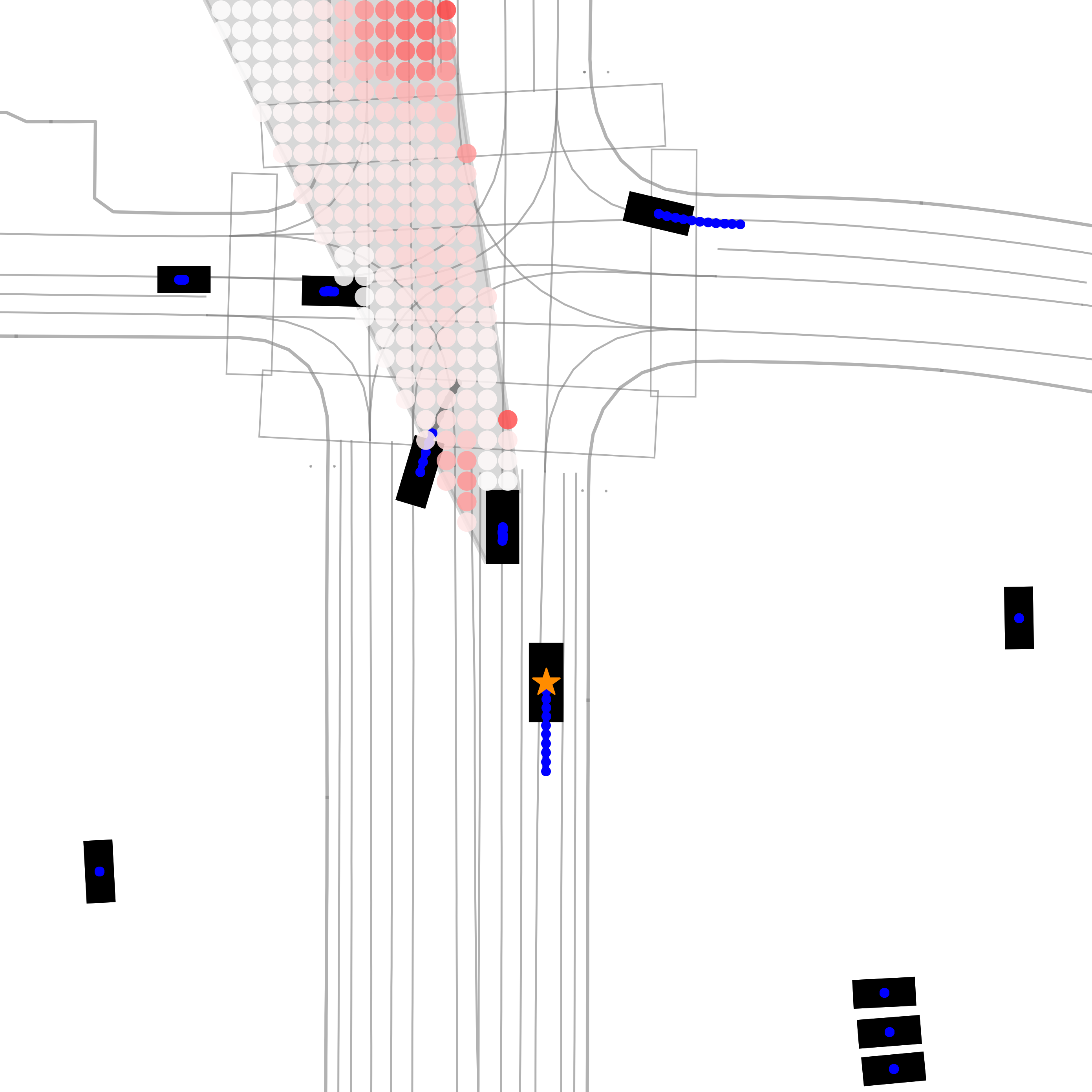}} 

\vspace{-0.5cm} 

\subfigure[Scene 1]{\label{fig:a}\includegraphics[width=36mm]{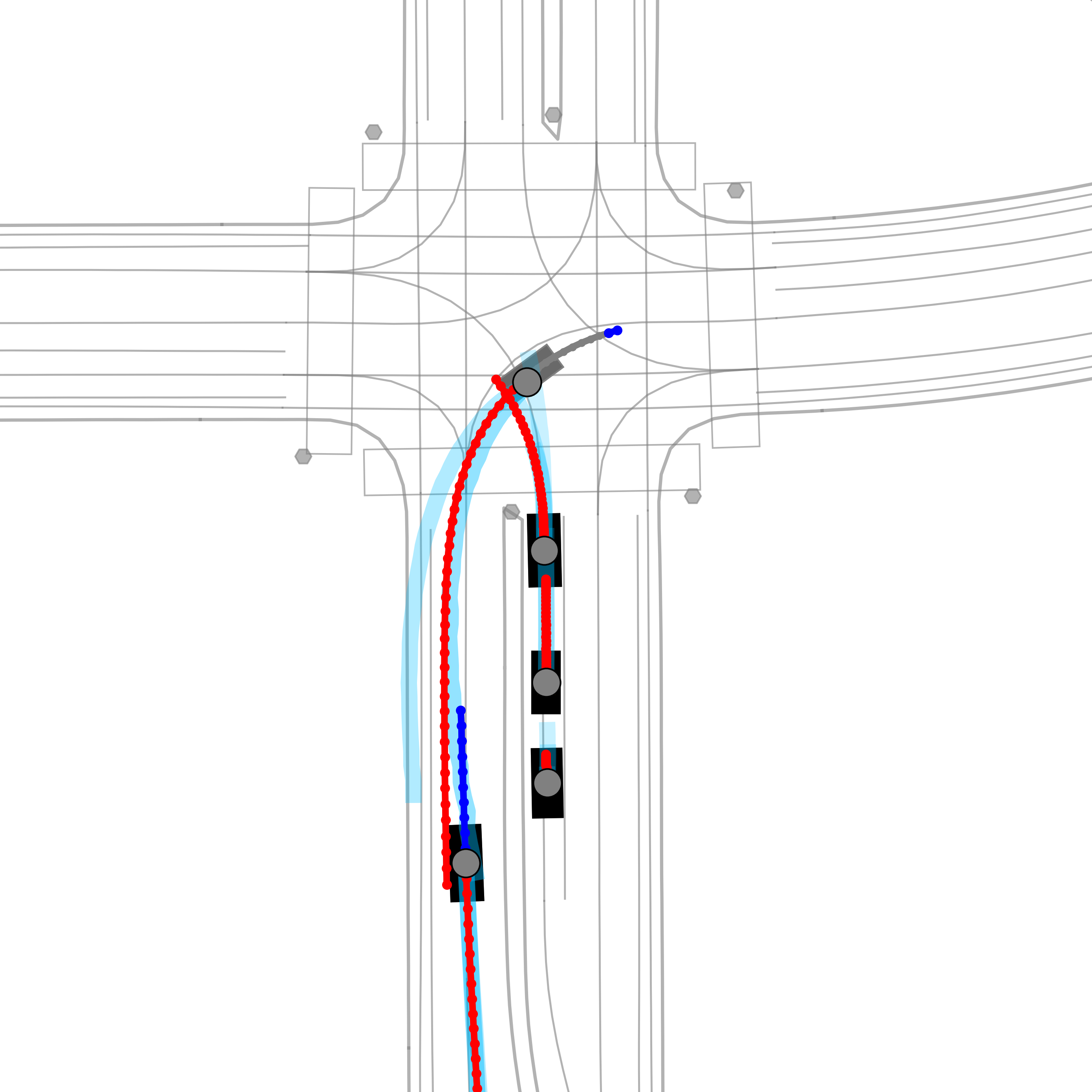}} 
\hspace{0.25cm} 
\subfigure[Scene 2]{\label{fig:b}\includegraphics[width=36mm]{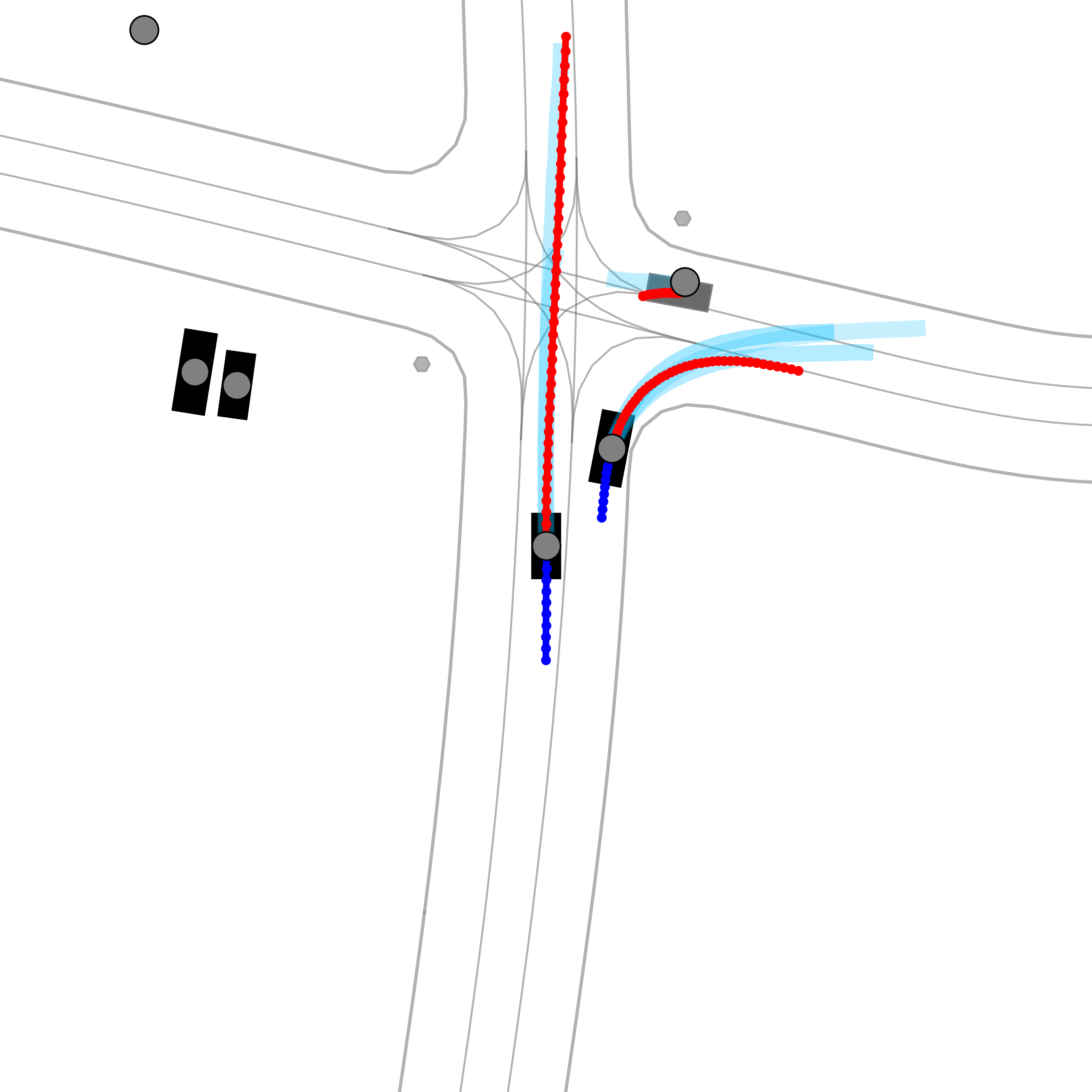}} 
\hspace{0.25cm} 
\subfigure[Scene 3]{\label{fig:a}\includegraphics[width=36mm]{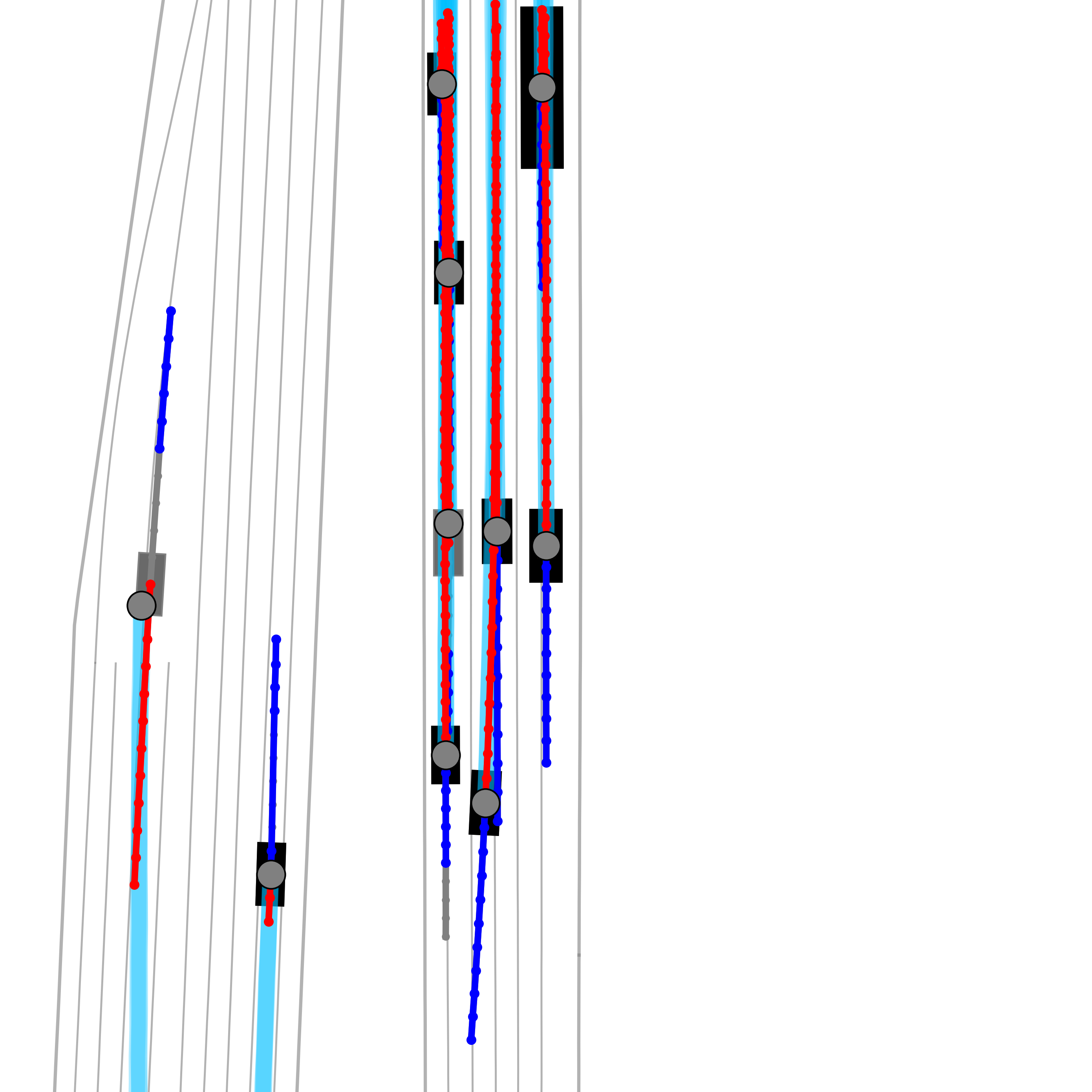}} 
\hspace{0.25cm} 
\subfigure[Scene 4]{\label{fig:b}\includegraphics[width=36mm]{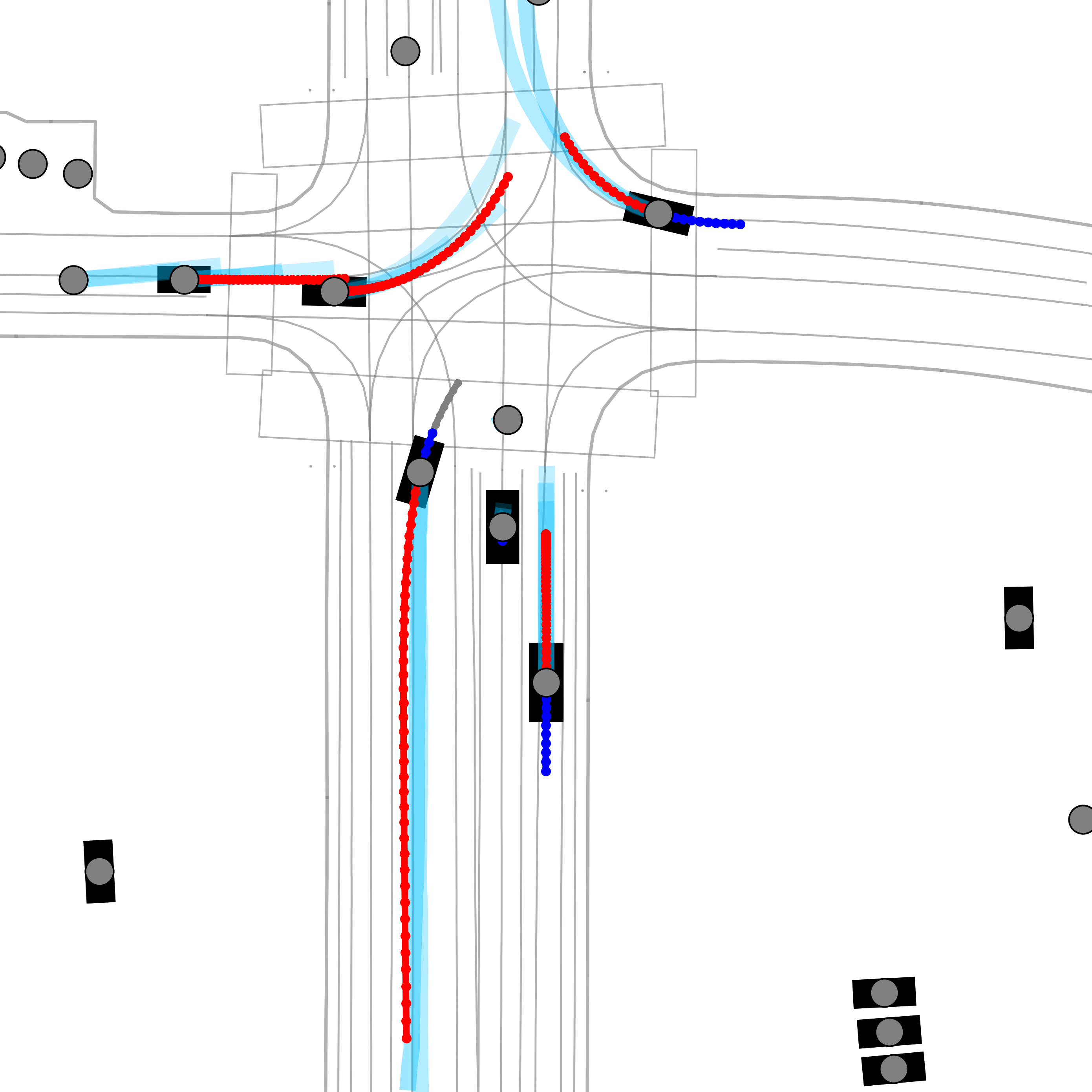}} 


\caption{Scene Informer Predictions in Crowded Settings: The \cyansquare\ visualizes predictions, its intensity reflecting trajectory probability. The \redanchor\ intensity signifies occupancy probability. The top row displays occupancy, the bottom, forecasted trajectories. The \greycircle\ denotes trajectories from high-likelihood occupancy anchors. Our method reliably predicts significant occupancy for ground truth occluded objects, delivering realistic trajectories for all agents. In scenes 1-3, our model accurately gauges occluded agent positions and their future paths. In scene 4, with the ego (\orangestar) nearing a crosswalk with stationary vehicles, our approach anticipates a crossing pedestrian.}
\vspace{-0.2cm}
\label{fig:results}
\end{figure*}

\renewcommand{\thesubfigure}{\alph{subfigure})}


The true AV's observability lies between full and limited observability. A suite of sensors combined with tracking systems can often detect what is behind obstructions and track agents through temporary occlusions. Nevertheless, an AV inherently lacks access to the true state of the environment, which underscores the importance of environment prediction in a partially observable setting. 

Each training sample consists of vectorized representations, such as \SI{1}{\second} of history, \SI{4}{\second} of ground truth future, polyline-based map representation, and a occlusion created by a single agent. Temporal information is discretized with \SI{0.1}{s} step. We reduce the dimensionality of the road map by sampling points separated by at least 1.5 meters and limiting the observable region to 60 meters around the ego vehicle. Anchors are assigned to the last time step of observed agents and randomly populated within the occlusion polygon. An example of a sample is shown in \cref{fig:dataset}. 

\subsection{Implementation details}

\textbf{Scene encoder:} We implement three MLPs for cars, bikes, pedestrians, and a polyline encoder for maps. Each output a set of feature vectors with a size of 256 for each observed agent and each polyline. We use a transformer encoder with four layers, four heads, 256 hidden size, and a 2048 feedforward dimension.

\textbf{Anchor decoder:} It consists of an MLP that encodes anchors, a two-layer transformer with other parameters the same as the encoder, and output MLP heads for occupancy classification, trajectory classification, and trajectory prediction. We consider seven future trajectory modes. 

\textbf{Training details:} The number of trainable parameters is 11.3M. We train each model with the AdamW~\cite{loshchilov2018decoupled} optimizer with a linear learning rate warmup from 0.0 to 0.0001 over the first 10k gradient steps. All models are implemented in PyTorch~\cite{paszke2019pytorch}, and trained with Lightning AI 2.0.6~\cite{Falcon_PyTorch_Lightning_2019} in mixed precision with a batch size of 20 accumulated over 2 steps for 10 epochs (250k gradient steps). Experiments were carried out on an NVIDIA TITAN RTX 24GB GPU with an AMD Ryzen 3960X CPU and 64 GB of RAM.

\subsection{Experimental Setup and Metrics}
We assess our framework's performance in terms of occlusion inference of unobserved agents and trajectory prediction of observed agents in a partially observable environment. 
Our evaluation metrics encompass the classification of occupancy for unobserved agents and a regression analysis comparing predicted trajectories to the ground truth for observed and unobserved agents. In the classification task, we assess the effectiveness in inferring which anchors are occupied within occlusions, focusing on the classification accuracy for occupied and free anchors ($ACC_{OCC}$/$ACC_{FREE}$). For the regression task, we evaluate the alignment between our predicted trajectories and the ground truth using minimum average and final displacement errors ($ADE_{min}$/$FDE_{min}$). 

We evaluate the performance on occluded agents by benchmarking against existing occlusion inference methods, such as K-means PaS~\citep{afolabi2018people}, GMM PaS~\citep{itkinaicra2022}, and MAVOI~\citep{itkinaicra2022}. None of the open-source approaches is capable of inferring future trajectories of occluded agents. To evaluate the inferred trajectory, we compare our method with variations of our framework: the vanilla trajectory prediction that reasons only about observed agents, and an occlusion inference adaptation that focuses exclusively on reasoning over occlusions. Even though the vanilla trajectory prediction assumes full observability, we can still query it to infer the trajectory of occluded agents by providing an anchor with the ground truth position of the occluded agent. 
For the observed agents, we compare Scene Informer with a vanilla trajectory prediction variations of our framework trained on fully observable data and limited observability data, respectively. The latter lets us determine whether we can enhance the robustness of vanilla trajectory prediction to partial observability without explicitly incorporating occlusion inference reasoning. 
We then assess the robustness of all models by evaluating them under various observability scenarios. This involves using a dataset where the likelihood of an occlusion being generated by an agent varies. Contrary to the training dataset, it applies to all agents potentially creating multiple occlusions in the scene. It ranges from \SI{0}{\percent} (full obs.), with increments of \SI{25}{\percent}, \SI{50}{\percent}, and \SI{75}{\percent} (partial obs.), to \SI{100}{\percent} (limited obs.) (see \cref{fig:dataset}).

\section{Discussion}
\label{sec:results}

\renewcommand{\thesubfigure}{}
\begin{figure*}[ht]
\tiny
\centering
\subfigure[Scene 1: Trajectories]{\label{fig:a}\includegraphics[width=33mm]{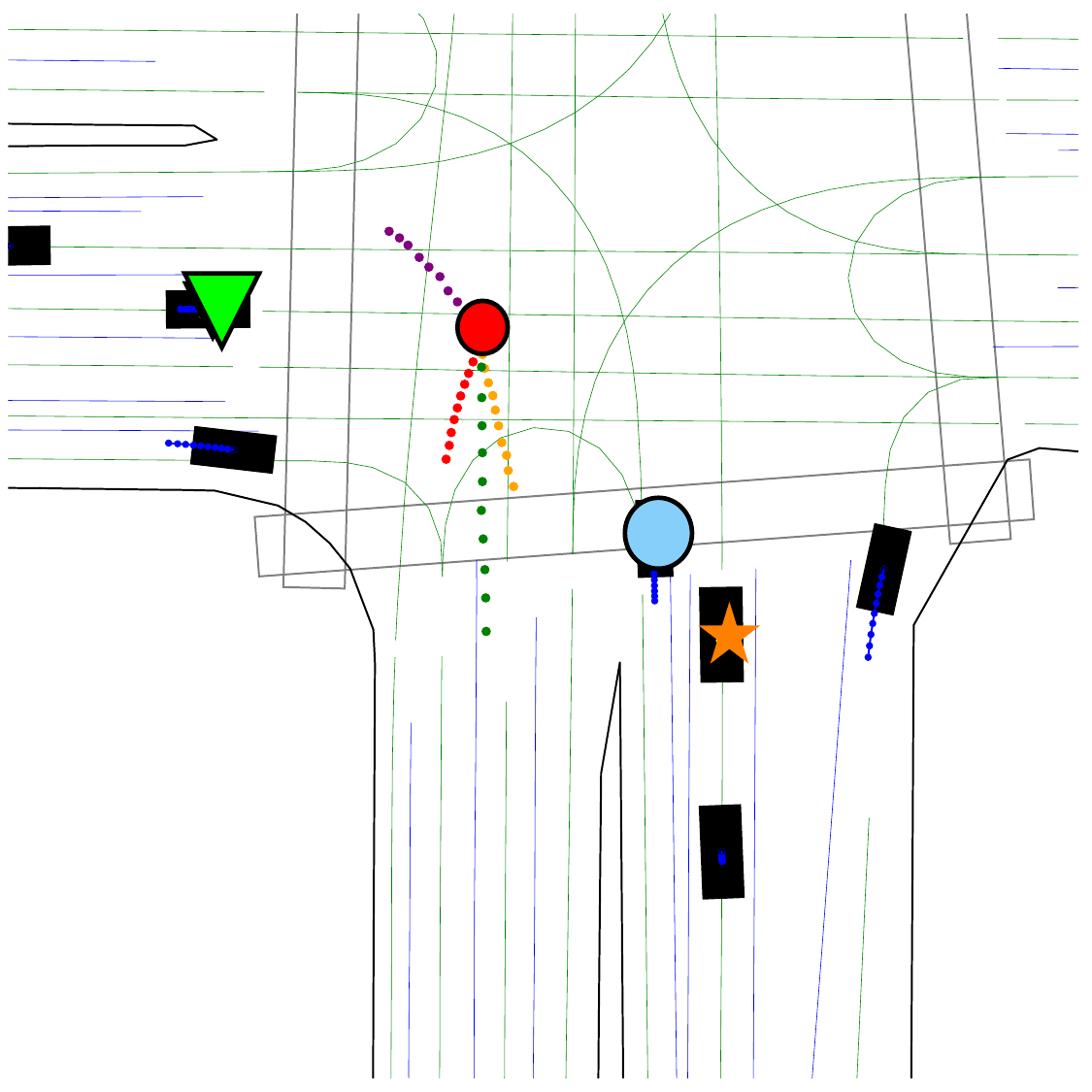}}
\hspace{1cm} 
\subfigure[Scene 1: Occupancy]{\label{fig:b}\includegraphics[width=33mm]{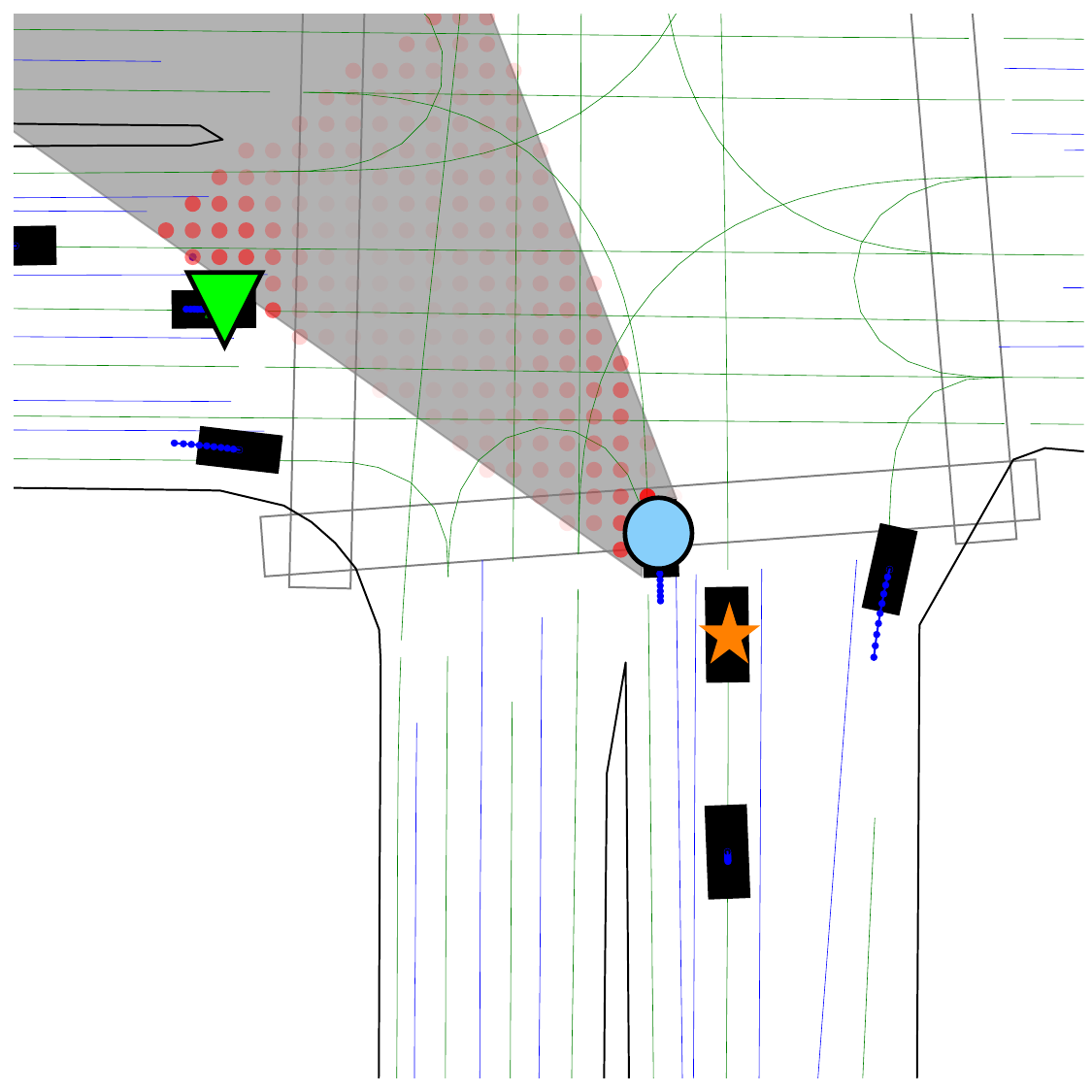}}
\hspace{1cm} 
\subfigure[Scene 2: Trajectories]{\label{fig:a}\includegraphics[width=33mm]{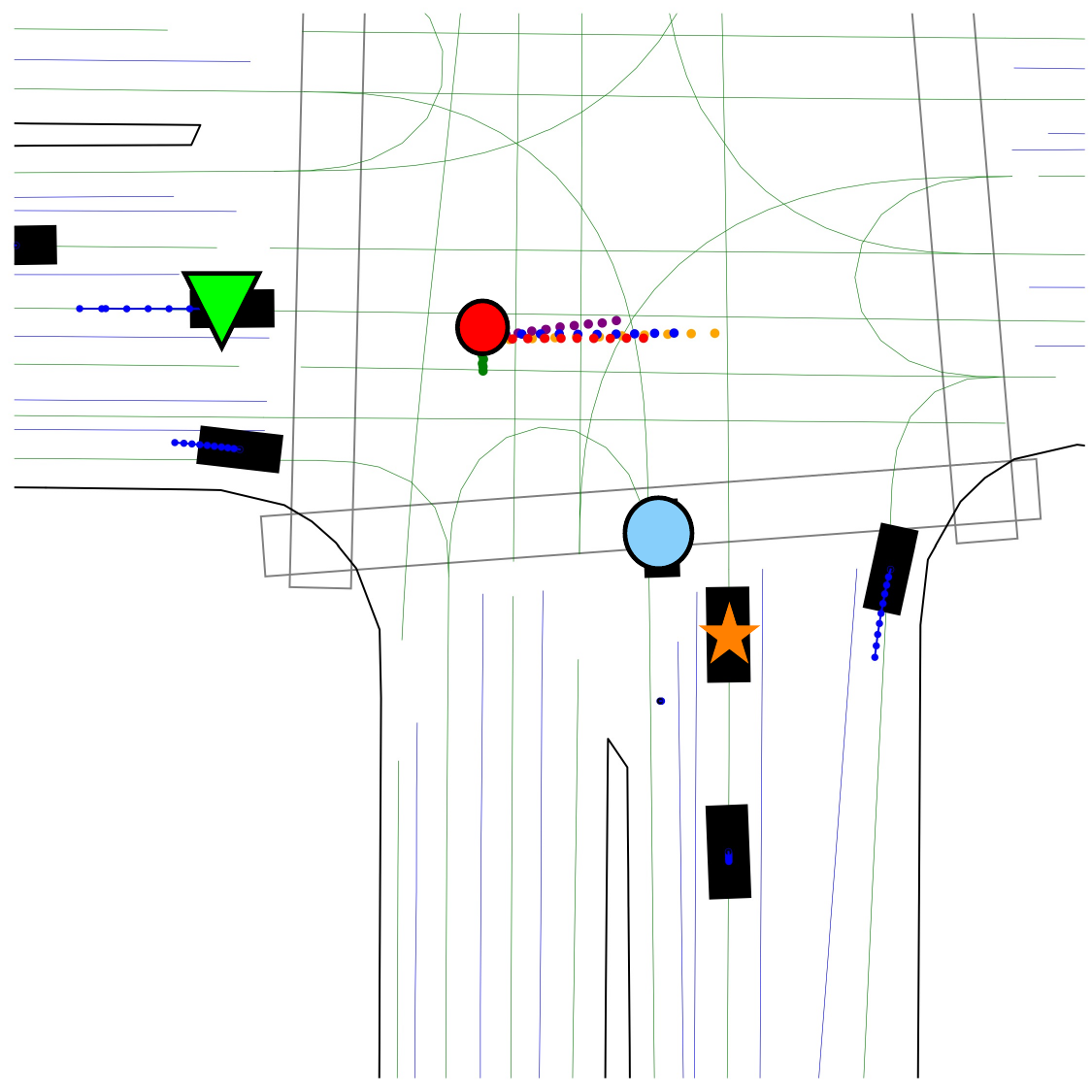}}
\hspace{1cm} 
\subfigure[Scene 2: Occupancy]{\label{fig:b}\includegraphics[width=33mm]{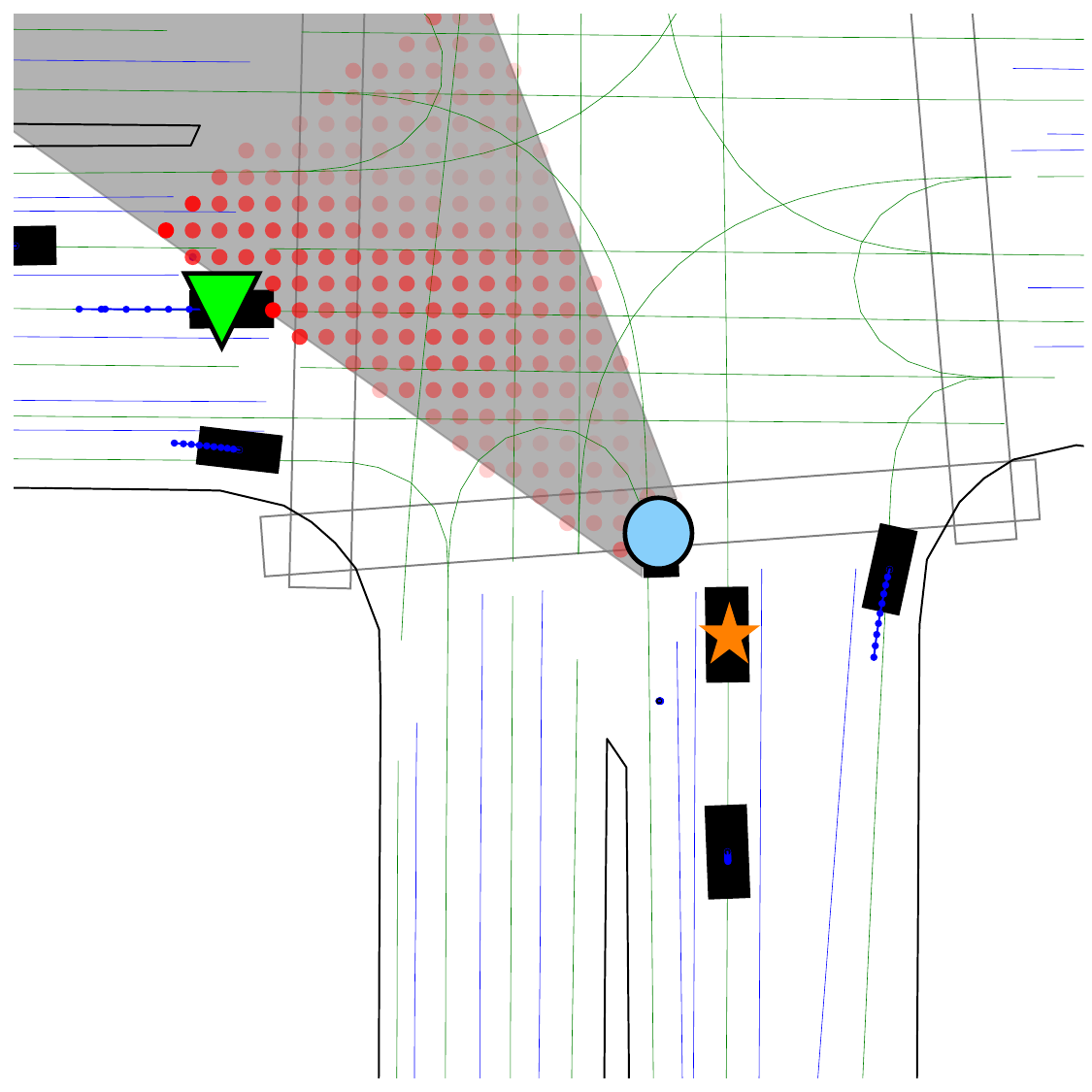}}
\caption{Impact of observed agents' histories on the occlusion inference performance. We modify the observed trajectories of two agents (\bluecircle and \greentriangle). In Scene 1, we visualize a scenario where a \greentriangle\ is stationary and a \bluecircle\ is moving forward. For an anchor \redanchor, our approach predicts occupancy with low probability and vertical motion from the top to the bottom of the intersection. In Scene 2, we visualize the predictions for a modified scenario. The \greentriangle\ is approaching quickly the intersection and the \bluecircle\ is stationary. Scene Informer realistically adapts its occupancy and trajectory predictions based on the observed motion of other agents. It is now highly likely that the anchor in the middle of the occlusion is occupied, and a majority of predicted trajectories are horizontal from the left to right of the intersection.}
  \label{fig:ablation}
 \end{figure*}
\renewcommand{\thesubfigure}{\alph{subfigure})}

\textbf{Occlusion inference performance}: 
We report the evaluation of the occlusion inference performance of our approach compared to the baselines and the impact of evaluation observability assumptions in \cref{table:results}. Our method consistently outperforms all baselines in both occupied and free accuracy metrics, regardless of the visibility setting. The improvement is in the range of \SI{10.5}{\percent}-\SI{27.5}{\percent} and \SI{17.0}{\percent}-\SI{35.0}{\percent} points on occupied and free cells, respectively. Notably, as the visibility assumption in the evaluation dataset gets more restrictive, our model tends to predict more occupied cells—a potentially desirable trait for safety-critical applications. 

On the regression task, Scene Informer outperforms a vanilla trajectory prediction baseline and a variation trained only on the occluded agents. As the observation becomes more limited, the performance gap between models, particularly when compared to vanilla trajectory prediction, widens in favour of our approach. In the limited observability setting, the $FDE_{min}$ of Scene Informer is \SI{2.37}{m} less than that of the vanilla trajectory prediction, which underscores the ability of our framework to robustly reason about occlusions across a range of observability scenarios. In addition, our framework outperforms its variation trained solely on occluded agents, which implies that occlusion inference and trajectory prediction are complementary tasks. 
 
\begin{table}[!t]
    \scriptsize
    \caption{Comparison on occlusion inference and trajectory prediction in terms of classification accuracy and displacement errors. For occluded agents, "Full" denotes "Full Obs (Occ. Inf.)".} 
    \setlength\tabcolsep{4pt} 
      \begin{center}
        \begin{tabular}{@{}lccccc@{}} \toprule
        Model & Full  & \multicolumn{3}{c}{Partial} & Limited \\ 
         & 0$\%$  & 25$\%$ & 50$\%$ & 75$\%$ & 100$\%$ \\ \midrule
         \multicolumn{6}{c}{Occlusion Classification Accuracy ($ACC_{OCC}(\uparrow)$\,/\,$ACC_{FREE}(\uparrow)$)}     \\  \midrule
        K-means PaS~\cite{afolabi2018people}  & 65.7\,/\,40.6 & 65.5\,/\,41.3  & 65.8\,/\,41.0 & 66.1\,/\,40.8 & 66.5\,/\,40.7\\
        GMM PaS~\cite{itkinaicra2022}  & 61.3\,/\,47.5 & 58.1\,/\,50.4 & 56.0\,/\,51.6 & 54.4\,/\,52.8 & 53.0\,/\,53.9\\
        MAVOI~\cite{itkinaicra2022}  & 67.9\,/\,58.6 & 65.5\,/\,55.7  & 67.1\,/\,56.1 & 67.3\,/\,55.4 & 66.7\,/\,55.0\\ 
        Ours & \textbf{78.4}\,/\,\textbf{75.6} & \textbf{79.8}\,/\,\textbf{74.0} & \textbf{80.4}\,/\,\textbf{73.4} & \textbf{80.7}\,/\,\textbf{72.6} & \textbf{80.5}\,/\,\textbf{72.8} \\ \midrule

        \multicolumn{6}{c}{Occluded Agents' Trajectory Prediction ($ADE_{min}(\downarrow)$/$FDE_{min}(\downarrow)$)}     \\  \midrule
        Traj. Pred.  & 1.94\,/\,2.97 & 2.08\,/\,3.22  & 2.23\,/\,3.49 & 2.37\,/\,3.76 & 2.50\,/\,4.00\\ 
        Occlusion Inf. Only  & 1.15\,/\,1.96  & 1.19\,/\,2.02 & 2.08\,/\,1.23 & 1.26\,/\,2.13 & 1.30\,/\,2.19\\
        Ours & \textbf{0.87}\,/\,\textbf{1.43} & \textbf{0.91}\,/\,\textbf{1.50} & \textbf{0.95}\,/\,\textbf{1.54} & \textbf{0.98}\,/\,\textbf{1.59} & \textbf{1.00}\,/\,\textbf{1.63} \\ \midrule
        \multicolumn{6}{c}{Observed Agents' Trajectory Prediction ($ADE_{min}(\downarrow)$/$FDE_{min}(\downarrow)$)}     \\  \midrule
        
        Traj. Pred. (Full Obs)  & \textbf{0.26}\,/\,\textbf{0.62} & 0.35\,/\,0.80  & 0.45\,/\,0.97 & 0.53\,/\,1.12 & 0.60\,/\,1.26\\
        Traj. Pred. (Limited)  & 0.28\,/\,0.67 & 0.33\,/\,0.78  & 0.38\,/\,0.88 & 0.42\,/\,0.97 & 0.45\,/\,1.04\\
        Ours (Full Obs) & \textbf{0.26}\,/\,\textbf{0.62} & \textbf{0.31}\,/\,\textbf{0.73} & \textbf{0.36}\,/\,\textbf{0.83} & \textbf{0.40}\,/\,\textbf{0.91} & \textbf{0.43}\,/\,\textbf{0.99} \\ \bottomrule
        \end{tabular}
        \vspace{-0.6cm}
        \label{table:results}
        \end{center}
\end{table} 

\Cref{fig:results} provides examples of inferred occlusions. As shown, our approach effectively predicts occupancy and the future trajectory based on other agents' motions and road layouts. The predicted occupancy map realistically reflects the true motion of partially observable traffic, infers potential unobserved agents and assigns a higher probability of occupancy to actively used parts of drivable spaces.
\Cref{fig:ablation} shows how the observed motion of the traffic participants impacts the occlusion inference. We modify the observations to indicate potentially different flow of traffic. We demonstrate that Scene Informer adapts its predictions realistically in response to changes in the behavior of observed agents. 

\textbf{Trajectory prediction performance}:
\Cref{table:results} compares the performance of our framework in predicting the trajectories of observed agents only and compares it with the commonly used vanilla trajectory prediction method. In a fully observable evaluation, our framework aligns with the performance of a vanilla trajectory prediction approach, as expected. However, as the evaluation accounts for increasing partial observability, our method consistently surpasses vanilla trajectory predictions, with the performance difference also becoming more evident with a higher probability of occlusions (\SI{0.27}{m} in final displacement error in the limited observability). Vanilla trajectory prediction methods assume that all interacting agents are present in the scene representations and do not account for missing or incomplete observations. 
The introduction of partial observability can then result in a corrupted scene representation, leading to inferior performance. In the real world, this manifests as flickering or missing agents due to physical obstructions or adverse weather conditions. Furthermore, a train-test distribution shift might arise due to differences in observation labeling strategies. While the training set may be manually labeled or automated by an off-board system, detections during deployment are provided by real-time, less accurate ones leading to incomplete observations. It underscores the importance of modeling partial observability and making trajectory prediction frameworks more robust.
We compare our approach with the vanilla trajectory prediction method, but trained in a limited observability setting. While it improves performance in the limited setting, it underperforms in the fully observable one. Yet, it still fails to match the performance of Scene Informer, which has been trained on the fully observable adaptation of the dataset. It indicates that incorporating occlusion reasoning enhances robustness in partial observability settings. Our frameworks can reason over agents that might be present but are not observed.
\section{CONCLUSION}
We present the first end-to-end comprehensive environment prediction framework that reasons about observable and unobservable parts of an environment. Scene Informer realistically infers any occlusion of interest and predicts future trajectories for observed agents. Unlike prior work in occlusion inference, it is an end-to-end framework and is not restricted to a fixed-size occupancy grid map. Moreover, it challenges the unrealistic full observability assumption in trajectory prediction. Our results demonstrate that Scene Informer surpasses existing occlusion inference methods, provides robustness in trajectory prediction to partial observability, and underscores the advantages of merging occlusion inference with trajectory prediction.
In future work, we aim to incorporate the interactions between occlusions, predict multiple possible occupancy probabilities, and explore its integration with planning. Many trajectory prediction models do not rely on any raw sensor data. This might cause them to overlook critical details~\cite{nayakanti2022wayformer, lange2022lopr}, leading to cascading errors~\cite{delecki2022we}. They also focus on trajectory predictions rather than joint scene predictions, which are critical for planning.

\section*{ACKNOWLEDGMENT}
This project was made possible by funding from the Ford-Stanford Alliance.

{\small
\printbibliography
}
\end{document}